\documentclass{article}

\usepackage{arxiv}

\usepackage[utf8]{inputenc} % allow utf-8 input
\usepackage[T1]{fontenc}    % use 8-bit T1 fonts
\usepackage{hyperref}       % hyperlinks
\usepackage{url}            % simple URL typesetting
\usepackage{booktabs}       % professional-quality tables
\usepackage{amsfonts}       % blackboard math symbols
\usepackage{nicefrac}       % compact symbols for 1/2, etc.
\usepackage{microtype}      % microtypography
\usepackage{lipsum}
\usepackage{graphicx}
\graphicspath{ {./images/} }

\usepackage{graphicx}
\usepackage{booktabs}
\usepackage[misc]{ifsym}

\usepackage{mwe}

\usepackage{microtype}
\usepackage{graphicx}
\usepackage{subfigure}
\usepackage{booktabs} % for professional tables
\usepackage{amssymb} 
\usepackage{amsmath}
\usepackage[section]{placeins}
\usepackage{algorithm}
\usepackage{algpseudocode}
\usepackage{makecell}

\usepackage{hyperref}
\hypersetup{hidelinks}

\usepackage{xcolor}
\usepackage{todonotes} % Für Inline-Kommentare

\definecolor{cGreen}{HTML}{4CAF50}
\definecolor{xRed}{HTML}{E53935}  % or 

\newcommand{\cmark}{\textcolor{cGreen}{\checkmark}}%
\newcommand{\xmark}{\textcolor{xRed}{\(\times\)}}%

\usepackage{amsmath}
\usepackage{amsfonts}
\usepackage{amssymb}
\usepackage{mathtools}
\DeclarePairedDelimiter{\ceil}{\lceil}{\rceil}

\newif \ifappx
\appxtrue

\title{Breaking Free: Decoupling Forced Systems with Laplace Neural Networks\thanks{Preprint. Accepted to the Research Track of ECML PKDD 2025. \textbf{Note:} This preprint corresponds to the originally submitted version prior to peer review and may differ from the Version of Record. The Version of Record will appear in the \textit{Proceedings of ECML PKDD 2025}, Lecture Notes in Artificial Intelligence (LNAI), Springer.}}

\author{
 Bernd Zimmering\\
  Institute for Artificial Intelligence\\
  Helmut Schmidt University\\
  Hamburg, Germany \\
  \texttt{bernd.zimmering@hsu-hh.de} \\
   \And
  Cecília Coelho \\
  Institute for Artificial Intelligence\\
  Helmut Schmidt University\\
  Hamburg, Germany \\
  \texttt{cecilia.coelho@hsu-hh.de} \\
  \And
 Vaibhav Gupta \\
   Institute for Artificial Intelligence\\
  Helmut Schmidt University\\
  Hamburg, Germany \\
  \texttt{guptav@hsu-hh.de} \\
  \AND
    Maria Maleshkova \\
      Institute for Artificial Intelligence\\
  Helmut Schmidt University\\
  Hamburg, Germany \\
   \texttt{maleshkm@hsu-hh.de} \\
   \And
   Oliver Niggemann \\
    Institute for Artificial Intelligence\\
  Helmut Schmidt University\\
  Hamburg, Germany \\
   \texttt{oliver.niggemann@hsu-hh.de} \\
}

\begin{document}
\maketitle
\begin{abstract}
Modelling forced dynamical systems—where an external input drives the system state—is critical across diverse domains such as engineering, finance, and the natural sciences. In this work, we propose Laplace-Net, a decoupled, solver-free neural framework for learning forced and delay-aware systems. It leverages a Laplace transform-based approach to decompose internal dynamics, external inputs, and initial values into established theoretical concepts, enhancing interpretability. Laplace-Net promotes transferability since the system can be rapidly re-trained or fine-tuned for new forcing signals, providing flexibility in applications ranging from controller adaptation to long-horizon forecasting. Experimental results on eight benchmark datasets—including linear, non-linear, and delayed systems—demonstrate the method’s improved accuracy and robustness compared to state-of-the-art approaches, particularly in handling complex and previously unseen inputs.
\end{abstract}

% keywords can be removed

\section{Introduction}
\label{sec:intro}

\begin{figure}[h]
    %\vspace{-0.15cm}
    \centering
    \includegraphics[width=0.4\linewidth]{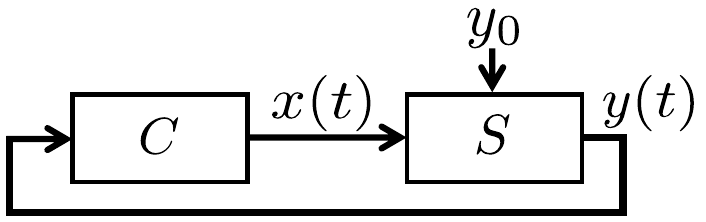}
    \caption{Control loop with system $S$ that responds with $y(t)$ and is forced by some controller $C$ through excitations $x(t)$. $y_0$ is the initial state of the system.}
    \label{fig:Forced_system.pdf}
\end{figure}

In \emph{forced} dynamical systems, an external controller $C$ (e.g., a motor or human) drives a dynamical system $S$ (e.g., a pendulum) using input signal $x(t)$ to achieve desired objectives over time $t$ (Fig.~\ref{fig:Forced_system.pdf}). The system responds with outputs $y(t)$, evolving from its initial state $y_0$ based on the interaction of the forcing inputs with its internal dynamics \cite{ogata2010}.
These systems span various domains such as control engineering, robotics, finance, ecosystems, and epidemiology, where external inputs (e.g., economic policies, climate, public health measures) strongly influence system behaviour \cite{ma2008hopf,valencia2015qualitative,dafilis2014dynamical}. Here, modelling $S$ is essential for designing $C$, enabling effective control of these complex systems.

Traditionally, modelling such systems has relied on hand-crafted ordinary differential equations (ODEs) with predefined forcing terms \cite{ogata2010}. These approaches require deep domain expertise and often struggle to generalise beyond their original design \cite{zolock2003application}.
 In recent years, data-driven methods like \emph{Neural ODEs} \cite{Chen2018NeuralOD} have emerged as powerful tools to learn the dynamics of the system directly from the data. Extensions of Neural ODEs \cite{Rubanova2019,kidger2020,gwak2020b,li24} incorporate external excitations $x(t)$ into their models, but rely on iterative ODE solvers. Such solvers can be computationally expensive and prone to numerical drift over long prediction horizons \cite{butcher1996}. Although alternative methods \cite{bilos2021,holt2022} have been proposed to address these challenges, they often fail to fully decouple the dynamics of the system $S$ from the external controller $C$, limiting their flexibility in handling arbitrary forcing inputs. 

 To overcome these limitations, \emph{Neural Operators} (NOs) have been introduced as a solver-free alternative for learning mappings directly from inputs $x(t)$ to outputs $y(t)$ \cite{lu2021,cao2024}. However, NOs frequently struggle to capture critical \emph{memory effects}, such as delays or non-local behaviours (i.e. fractional differential equations (FDEs) as described in \cite{diethelm2002b}), which are essential for accurately modelling many real-world systems. Recent studies \cite{zhu2021a,coelho2024a,pmlr-v255-monsel24a} emphasise the importance of addressing these memory effects to improve model fidelity.

Another promising avenue involves leveraging the Laplace Domain (LD) for learning forced dynamical systems. LD-based methods have shown strong performance in capturing both forced dynamics and delays \cite{cao2024,holt2022}. Although the Laplace transform has long been a cornerstone in fields like control theory, fluid dynamics, and systems biology, its integration into modern machine learning frameworks remains relatively under-explored.

In many real-world scenarios - such as adapting robotic controllers to new hardware configurations or refining economic forecasts under changing regulations - modular and \emph{decoupled} architectures offer significant advantages. By isolating specific components of a model (e.g., internal dynamics vs. external forcing), these architectures enable efficient adaptation to new scenarios without requiring a complete model retraining. This modularity not only reduces computational costs but also enhances interpretability by aligning learned components with classical theoretical insights.

Despite the progress made by existing methods, many remain constrained by specific assumptions about input signals, overlook delays or other memory effects, or rely heavily on iterative solvers. 
These limitations can lead to numerical drift, high computational overheads, or inadequate modelling of long-term dependencies - especially in scenarios involving complex forcing signals or extended prediction horizons.

 To address these challenges, we propose a novel modular Laplace-based framework that explicitly handles external forcing and memory effects without relying on iterative solvers (as summarised in Table \ref{tab:comparison}). Our main contributions can be summarized as follows:
\begin{itemize}
    \item \textbf{Decoupled Laplace Representation:} We introduce LP-Net, a decoupled NN architecture that employs the Laplace domain with an explicit factorization that separates internal system dynamics from external forcing and initial states, aligning with classical system theory.
    \item \textbf{Arbitrary Forcing and Intervention Handling:} Our approach accepts time-varying or previously unseen inputs without retraining the entire model, facilitating straightforward adaptation to new control signals or external perturbations.
    \item \textbf{Solver-Free, Memory-Aware Inference:} We follow Holt et al. \cite{holt2022} employing numerical inverse Laplace transforms and thus avoiding iterative integration. We mitigate numerical drift and capture memory effects (e.g., delays or fractional dynamics) within the same framework.
    \item \textbf{Enhanced Transferability and Interpretability:} The decoupling into well established subcomponents enables their reuse or pre-training and improves interpretability for experts.
    \item \textbf{Improved Performance:} LP-Net consistently outperforms LNO across all datasets and surpasses LSTM on 6 out of 8 datasets, demonstrating its effectiveness in capturing linear, non-linear, chaotic, and memory-dependent dynamics.
 
\end{itemize}

This remainder of this paper is structured as follows:
Section~\ref{sec:Preliminaries} formally defines the problem and introduces key concepts related to Laplace transforms. Section~\ref{sec:Solution} presents the proposed decoupled Laplace-based framework. Section~\ref{sec:experiments} evaluates our approach on linear, non-linear and delayed systems.\footnote{Code is available at \url{https://github.com/zimmer-ing/Laplace-Net}} The conclusions and future directions are discussed in Section~\ref{sec:Discussion}.

\section{Related Work}
\label{sec:relatedwork}

Data-driven modelling of dynamical systems encompasses a wide range of neural approaches, many of which differ in their reliance on time-stepping solvers, ability to model memory effects or inhomogeneous excitations, and degree of modular decomposition. 
Table~\ref{tab:comparison} summarizes representative methods evaluated across four key aspects: 
(i) \emph{Solver-Free} indicates whether the method avoids sequential time-stepping during training; 
(ii) \emph{Memory} assesses the capability to explicitly handle delays or non-local behaviour; 
(iii) \emph{Arbitrary Forcing} evaluates the ability to generalize to unseen or non-parametric input trajectories; 
(iv) \emph{Decoupled} reflects the extent to which the approach separates system dynamics, external inputs, and initial conditions.

\begin{table*}[ht]
\centering
\caption{Comparison of neural approaches for modelling dynamical systems along four key dimensions: solver-free inference; memory (e.g. delays) ; arbitrary forcing; modularity.}
\label{tab:comparison}

\begin{tabular}{lccccc}
\toprule
\textbf{Method} & \textbf{Reference} & \shortstack{\textbf{Solver} \\ \textbf{Free}} & \textbf{Memory} & \shortstack{\textbf{Arbitrary} \\ \textbf{Forcing}} & \textbf{Decoupled}  \\
\midrule
Neural ODE      &\cite{Chen2018NeuralOD}& \xmark & \xmark & \xmark & \xmark \\
Neural FDE      &\cite{coelho2024a,pmlr-v255-zimmering24a}& \xmark & \cmark & \xmark & \xmark\\
Neural DDE      &\cite{zhu2021a}               & \xmark & \cmark & \xmark & \xmark \\
ODE-RNN         &\cite{Rubanova2019}           & \xmark & \xmark & \cmark & \xmark \\
Fourier NODE    &\cite{li24}                   & \xmark$^\dagger$ & \xmark & \xmark & \xmark \\ 
Neural IM       &\cite{gwak2020b}              & \xmark & \xmark & \cmark & \cmark \\
Neural CDE      &\cite{kidger2020}             & \xmark & \xmark & \cmark & \xmark\\
Neural Flow     &\cite{bilos2021}              & \cmark & \xmark & \xmark & \xmark \\
Neural Laplace  &\cite{holt2022}               & \cmark & \cmark & \xmark & \xmark \\
DeepONet        &\cite{lu2021}                 & \cmark & \xmark & \cmark & \xmark$^\ddagger$\\
Laplace NO      &\cite{cao2024}                & \cmark & \xmark & \cmark & \cmark \\
\midrule
\textbf{LP-Net}& This Work                  & \cmark & \cmark & \cmark & \cmark\\
\bottomrule
\end{tabular}
\begin{flushleft}
\small $^\dagger$ Fourier NODEs eliminate the need for a solver during training but still require one for inference.\\
\small $^\ddagger$ DeepONet is partially decoupled: the branch network encodes both initial and external inputs, while the trunk network models system dynamics. However, the separation between initial conditions and external inputs within the branch network is not complete.
\end{flushleft}
\end{table*}

\paragraph{\bf Neural ODEs and Solver-Based Extensions.}
\emph{Neural ODEs}~\cite{Chen2018NeuralOD} and their immediate variants, such as \emph{ODE-RNN}~\cite{Rubanova2019}, \emph{ANODE}~\cite{dupont2019a}, and \emph{Neural IM}~\cite{gwak2020b}, parameterise ODE vector fields in a continuous latent space. These approaches have demonstrated their effectiveness in various scenarios, including homogeneous and some forced systems. However, they still rely on iterative numerical solvers such as Runge-Kutta, which can lead to substantial computational costs during training and error accumulation over extended time horizons.
Extensions of this framework include approaches for delayed~\cite{zhu2021a,pmlr-v255-monsel24a} and fractional systems. \emph{Neural Delayed DEs}~\cite{zhu2021a} and \emph{Neural Fractional DEs}~\cite{coelho2024b,pmlr-v255-zimmering24a} explicitly embed delay or learn the amount of memory \cite{coelho2024a} but, like their predecessors, rely on solver-based training loops and do not explicitly separate forcing inputs.

\paragraph{\bf Laplace-Based, Solver-Free Methods.}
\emph{Neural Laplace} \cite{holt2022} revolutionizes the approach to learning DEs by operating in the Laplace domain, eliminating the need for stepwise integration during both training and prediction. This method employs a numerical inverse Laplace transform algorithm to generate time-domain predictions, enabling the capture of complex memory-like effects such as fractional-order behaviour and delay times. While \emph{Neural Laplace} claims to handle forced differential equations, its implementation is limited to forcing functions that remain constant between training and testing phases, rather than accommodating arbitrary inputs.
Extensions \cite{pmlr-v206-holt23a} adapt the framework for reinforcement learning scenarios. However, it only considers past actions during prediction, not accounting for future actions, restricting its applicability.
\emph{Fourier NODE}~\cite{li24} offers an alternative approach, approximating state derivatives in the frequency domain to avoid using ODE solvers during training. This method explicitly incorporates control inputs for trajectory prediction, enhancing its versatility in handling forced systems. However, \emph{Fourier NODE} employs a solver-free approach during training but still requires a numerical solver for making predictions during inference.

\paragraph{\bf Operator Learning for Forced Systems.}
Modern approaches, such as \emph{DeepONets}~\cite{lu2021} and \emph{Laplace Neural Operators}~\cite{cao2024}, enable \emph{solver-free} evaluations by learning mappings from input (forcing) functions directly to solution functions. While initially developed for Partial Differential Equations (PDE), some works have successfully extended their application to ODEs \cite{lu2021,cao2024}.
\emph{DeepONets}, for instance, employ a unique architecture consisting of two key components: a branch network that encodes inputs functions, such as initial conditions and forcing profiles, into a high-dimensional vector; a trunk network that processes evaluation points of the solution. This structure allows \emph{DeepONets} to effectively capture complex solution behaviours across a wide range of input conditions. However, the method's unified approach to encoding both system dynamics and external influences can limit the interpretability and modular reuse of the learned representations.
In contrast, \emph{Laplace Neural Operators} introduce a degree of modularization by separating internal system dynamics from external inputs within the Laplace domain. By restricting themselves to a pole--residue representation (Eq.~(5) in \cite{cao2024}), these operators enable the inverse Laplace transform to be carried out symbolically, simplifying implementation. However, such a representation struggles to capture delayed or fractional dynamics (Eq.~(15) in \cite{kexue2011}, Theorem~5.4 in \cite{Chen2015LAPLACETO}, and Example~1.25 in \cite{schiff1999}), as their Laplace transforms do not generally reduce to simple pole--residue forms. Methods like \emph{Neural Laplace}~\cite{holt2022} address this limitation by numerically computing the inverse Laplace transform, thereby allowing for a wider range of dynamical phenomena beyond the pole--residue framework.

\paragraph{\bf Bridging Gaps via Decoupling and Generalized Forcing.}
As highlighted in Table~\ref{tab:comparison}, a recurring shortfall in existing methods is the lack of an \emph{explicit} factorization of the system’s internal transfer characteristics, initial conditions, and arbitrary forcing signals — especially in a solver-free framework. Some methods, such as \emph{Neural IM}~\cite{gwak2020b} and \emph{ODE-RNN}~\cite{Rubanova2019}, partially decouple interventions or control inputs, but they often rely on iterative updates over time or fail to handle continuous forcing seamlessly. Similarly, \emph{Neural Laplace Control}~\cite{pmlr-v206-holt23a} extends Holt et al. \cite{holt2022} by incorporating past actions into a latent state representation, yet it assumes a homogeneous response for predictions and does not account for future forcing inputs, limiting its applicability in scenarios requiring forward-looking control.

\section{Preliminaries}
\label{sec:Preliminaries}
We consider a dynamical system \(\mathcal{S}\) with input \(\mathbf{x}(t) \in \mathbb{R}^{D_x}\) and response \(\mathbf{y}(t) \in \mathbb{R}^{D_y}\).  
Let \(\mathbf{t}=(t_1,\dots,t_{N+M})\subset[0,T]\) be discrete times (\(t_1 < \cdots < t_{N+M}\)), where the first \(N\) indices partition the \emph{historical} segment $\mathbf{t}_\text{hist}$ and the final \((N+1,\dots,N+M)\) indices partition the \emph{forecast} segment $\mathbf{t}_\text{fore}$. We collect samples into \(\mathbf{X} \in \mathbb{R}^{(N+M)\times {D_x}}\) and \(\mathbf{Y} \in \mathbb{R}^{(N+M)\times {D_y}}\)
with the partitions \(\mathbf{X}_{\mathrm{hist}}\in \mathbb{R}^{N\times {D_x}}\), \(\mathbf{X}_{\mathrm{fore}}\in \mathbb{R}^{M\times p}\), \(\mathbf{Y}_{\mathrm{hist}}\in \mathbb{R}^{N\times q}\), and \(\mathbf{Y}_{\mathrm{fore}}\in \mathbb{R}^{M\times q}\) defined analogously. Our objective is to predict \(\mathbf{Y}_{\mathrm{fore}}\) given \(\mathbf{X}_{\mathrm{hist}}, \mathbf{Y}_{\mathrm{hist}}, \) and \(\mathbf{X}_{\mathrm{fore}}\). Formally, we learn
\begin{equation}
\label{eq:task}
    f: \bigl(\mathbf{X}_{\mathrm{hist}},\,\mathbf{Y}_{\mathrm{hist}},\,\mathbf{X}_{\mathrm{fore}}\bigr)
    \,\mapsto\, \mathbf{Y}_{\mathrm{fore}}.
\end{equation}

Although the above forecasting objective is stated in the time domain, a powerful way to analyse and solve differential equations is via the Laplace transform. We therefore briefly recall the key properties of this transform and its inverse as they form the foundation of our solution.

\paragraph{The Laplace transform} maps time-domain signals into the complex \(s\)-domain, where derivatives become algebraic factors. For a function \(\mathbf{y}(t)\), the Laplace transform is:  
\begin{equation}
    \label{eq:LT_analytic}
    \boldsymbol{\mathcal{Y}}(s) \;=\; \mathcal{L}\{\mathbf{y}(t)\} 
  \;=\; \int_{0}^{\infty} e^{-st}\,\mathbf{y}(t)\,\mathrm{d}t, \quad s \in \mathbb{C}.
\end{equation}
The complex variable \( s \) is typically written as \( s = \sigma + i\omega \), where \( \sigma \in \mathbb{R} \) corresponds to exponential growth/decay and \(\omega \in \mathbb{R}\) to oscillatory behaviour. Applied to the \(n\)-th derivative, the Laplace transform yields:
\begin{equation}
\label{eq:LP_differential}
    \mathcal{L}\!\Bigl\{\frac{d^n\mathbf{y}(t)}{dt^n}\Bigr\}
\;=\;
s^n\,\boldsymbol{\mathcal{Y}}(s)
\;-\;
\sum_{k=0}^{\,n-1}
s^{\,n-1-k}\;
\frac{d^k\mathbf{y}(0)}{dt^k}.
\end{equation}
A pure time delay \(\tau\) appears as \(e^{-\tau s}\) in the Laplace domain, and fractional derivatives can be treated similarly \cite{schiff1999,Chen2015LAPLACETO,kexue2011}.

Solving an ODE in the Laplace domain calls for the inverse Laplace transform (ILT),
\begin{equation}\label{eq:ILT_analytic}
  \mathbf{y}(t) \;=\; \mathcal{L}^{-1}\{\boldsymbol{\mathcal{Y}}(s)\} 
  \;=\; \frac{1}{2\pi i}\,\int_{\sigma - i\infty}^{\sigma + i\infty} \boldsymbol{\mathcal{Y}}(s)\,e^{st}\,\mathrm{d}s,
\end{equation}
which is rarely solvable in closed form. 

Since direct evaluation of Eq. \eqref{eq:ILT_analytic} is rarely feasible, numerical methods are required. Among them, the Fourier series-based ILT \cite{dubner1968} was identified as the most robust for boundary element simulations \cite{kuhlman2013} and proposed for machine learning due to its efficiency and stability \cite{holt2022}.
 This method reconstructs smooth, real-valued signals by sampling \(\boldsymbol{\mathcal{Y}}(s)\) along a shifted vertical contour in the complex plane:
\begin{equation}\label{eq:ILT_Fourier}
    \mathbf{y}(t) \approx \frac{1}{\lambda} e^{\sigma t} \Biggl[
    \frac{\boldsymbol{\mathcal{Y}}(s_0)}{2} 
    + \sum_{k=1}^{N_{\text{ILT}}} \mathrm{Re} \Bigl\{ \boldsymbol{\mathcal{Y}}(s_k)\,e^{\,i \frac{k\pi t}{\lambda}} \Bigr\}
    \Biggr].
\end{equation}

Here, \(\lambda\) controls the frequency resolution, \(N_{\text{ILT}}\) determines accuracy, and \(s_0, s_k\) are the query points given by Eq.~\eqref{eq:queries}. The parameter \(\sigma = \alpha - \frac{\log(\epsilon)}{\lambda}\) shifts the contour to the right of all singularities, ensuring numerical stability. 

In practice, choosing \(\lambda\) proportional to $t$ mitigates numerical instabilities caused by the exponential factor. Setting $\lambda = \zeta t $ for $ t > 0$ results in time dependent query points:
\begin{equation}
\label{eq:queries}
    s_k(t) = \sigma + i \frac{k\pi}{\zeta t}, \quad k \in \mathbb{N}_0, \quad k \leq N_{\text{ILT}}, \quad \text{with} \quad \sigma = \alpha - \frac{\log(\epsilon)}{\zeta t}.
\end{equation}

This structured approach ensures that the contour remains stable while maintaining high accuracy in the reconstruction.

\section{Solution}
\label{sec:Solution}
To model Eq. \eqref{eq:task}, we propose LP-Net, a Laplace-based neural network (NN) that removes the solver, decouples initial conditions from inputs, and encapsulates system dynamics. We first decompose the system response in line with classical control theory \cite{ogata2010} and then integrate NNs into this framework.
\subsection{Decomposition of System Responses}

To illustrate the decomposition of \(\boldsymbol{\mathcal{Y}}(s)\) into decoupled components, we consider a differential equation of the form:
\begin{equation}
\label{eq:ODE_multi}
    \sum_{i=0}^{n} \mathbf{A}_i \,\frac{d^i\mathbf{y}(t)}{dt^i} 
    \;=\; 
    \mathbf{B}\,\mathbf{x}(t),
\end{equation}
where \(\mathbf{A}_i \in \mathbb{R}^{D_y \times D_y}\) and  
\(\mathbf{B} \in \mathbb{R}^{D_y\times D_x}\) are constant matrices.  

Applying the Laplace transform \(\mathcal{L}\{\cdot\}\) to both sides and using Eq. \eqref{eq:LP_differential} yields:
\begin{equation}
\label{eq:ODE_multi_LP}
    \sum_{i=0}^{n}
    \mathbf{A}_i
    \biggl(
        s^i\,\boldsymbol{\mathcal{Y}}(s)
        \;-\;
        \sum_{k=0}^{i-1} s^{i-1-k}\,\frac{d^k \mathbf{y}(0)}{dt^k}
    \biggr)
    =\mathbf{B}\boldsymbol{\mathcal{X}}(s).
\end{equation}
Rearranging terms and isolating \(\boldsymbol{\mathcal{Y}}(s)\), assuming the invertibility of  
\(\Bigl(\sum_{i=0}^{n} \mathbf{A}_i s^i \Bigr)^{-1}\), results in:
\begin{equation}
\label{eq:ODE_multi_LP_Y}
\boldsymbol{\mathcal{Y}}(s)
=
\Bigl(\sum_{i=0}^{n} \mathbf{A}_i s^i \Bigr)^{-1}
\Biggl(
\mathbf{B}\boldsymbol{\mathcal{X}}(s)
+
\sum_{i=0}^{n} \mathbf{A}_i \sum_{k=0}^{i-1} s^{i-1-k} \frac{d^k \mathbf{y}(0)}{dt^k}
\Biggr).
\end{equation}
o simplify the structure of the solution, we introduce the notation  
$\boldsymbol{\mathcal{H}}(s) \in \mathbb{C}^{D_y \times D_y}$ and  
$\boldsymbol{\mathcal{P}}(s) \in \mathbb{C}^{D_y}$ as follows:
\begin{equation}
\label{eq:transfer_function}
    \boldsymbol{\mathcal{H}}(s) := \Bigl(\sum_{i=0}^{n} \mathbf{A}_i s^i \Bigr)^{-1}, 
    \quad \boldsymbol{\mathcal{H}}(s) \in \mathbb{C}^{D_y \times D_y},
\end{equation}
\begin{equation}
\label{eq:P_func}
    \boldsymbol{\mathcal{P}}(s) := \sum_{i=0}^{n} \mathbf{A}_i \sum_{k=0}^{i-1} s^{i-1-k} \frac{d^k \mathbf{y}(0)}{dt^k}, 
    \quad \boldsymbol{\mathcal{P}}(s) \in \mathbb{C}^{D_y}.
\end{equation}
Using Eqs. \eqref{eq:transfer_function} and \eqref{eq:P_func} we can simplify Eq. \eqref{eq:ODE_multi_LP_Y} to:
\begin{equation}
\label{eq:decomposed_LP}
    \boldsymbol{\mathcal{Y}}(s) = \boldsymbol{\mathcal{H}}(s) \bigl(\mathbf{B}\boldsymbol{\mathcal{X}}(s) + \boldsymbol{\mathcal{P}}(s)\bigr),
\end{equation}
which represents the decomposition of Eq. \ref{eq:ODE_multi} into separate components: 
the system dynamics given by $\boldsymbol{\mathcal{H}}(s)$, 
the influence of initial conditions captured in $\boldsymbol{\mathcal{P}}(s)$, 
and the external excitations $\mathbf{B}\boldsymbol{\mathcal{X}}(s)$.

\subsection{Neural Network-based Approximation}  
\begin{figure}[h]
    \centering
    \includegraphics[width=0.8\columnwidth]{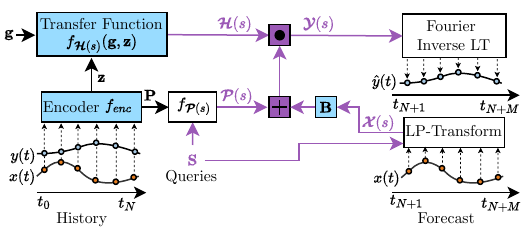}
    \caption{Overview of the LP-Net architecture. \textcolor[HTML]{3399CC}{Blue} elements represent learnable matrices or NNs, including an encoder for historical data and a trainable transfer function. \textcolor[HTML]{9B59B6}{Purple} elements denote complex-valued components.}
    \label{fig:LPNet}
\end{figure}

To generalize the decomposition in Eq.~\eqref{eq:decomposed_LP}, we introduce \textbf{LP-Net}. Figure~\ref{fig:LPNet} provides an overview, and Algorithm~\ref{alg:lpnet} details the computational steps. First, historical input-output sequences are encoded into $(\mathbf{P}, \mathbf{z})$, where $\mathbf{P}$ captures initial conditions. Using $\mathbf{P}$ and queries $\mathbf{s}$, $\boldsymbol{\mathcal{P}}(s)$ is formed in the Laplace domain. Second, the external input $\mathbf{X}_\text{fore}$ undergoes a numerical Laplace transform and is mapped into the output space via $\mathbf{B}$. Third, a NN approximates the transfer function $\boldsymbol{\mathcal{H}}(s)$, which, together with the other components, is combined using complex-valued operations to compute $\boldsymbol{\mathcal{Y}}(s)$ in the Laplace domain. Finally, an inverse Laplace transform reconstructs the time-domain output $\mathbf{Y}_\text{fore}$. To handle long sequences and non-linearities, this process runs recurrently with an adjustable stride $\delta$, resulting in $Q$ steps.  

\begin{algorithm}
    \caption{Neural Laplace (LP-Net)}
    \label{alg:lpnet}
    \begin{algorithmic}[1]
        \Require $(\mathbf{t}_\text{hist}, \mathbf{X}_\text{hist}, \mathbf{Y}_\text{hist})$, $(\mathbf{t}_\text{fore}, \mathbf{X}_\text{fore})$, $\delta$, $N_{\mathrm{ILT}}$
        \Ensure $\mathbf{Y}_\text{pred}$
        
        \Function{LPNet}{$\mathbf{t}_\text{hist}, \mathbf{X}_\text{hist}, \mathbf{Y}_\text{hist}, \mathbf{t}_\text{fore}^{(z)}, \mathbf{X}_\text{fore}^{(z)}, N_{\mathrm{ILT}}$}
            \State $(\mathbf{P}, \mathbf{z}) \gets f_{\mathrm{enc}}(\mathbf{t}_\text{hist}, \mathbf{X}_\text{hist}, \mathbf{Y}_\text{hist})$
            \State Compute Queries $\mathbf{s}$ from $\mathbf{t}_\text{fore}^{(z)}$ by Eq. \eqref{eq:queries}
            \State $\boldsymbol{\mathcal{P}}(s) \gets f_{\boldsymbol{\mathcal{P}}(s)}(\mathbf{P},\mathbf{s})$  \Comment{Eq. \eqref{eq:init_Ps}}
            \State Compute grid $\mathbf{g} $ from $\mathbf{t}_\text{fore}$,$N_{ILT}$ by Eq. \eqref{eq:grid}
            \State $\boldsymbol{\mathcal{H}}(s) \gets f_{\boldsymbol{\mathcal{H}}(s)}(\mathbf{g}, \mathbf{z})$\Comment{Eq.\eqref{eq:Hs_net}}
            \State $\boldsymbol{\mathcal{H}}(s) \gets \frac{\boldsymbol{\mathcal{H}}(s)}{f_{\mathrm{scale}} (\mathbf{t}_\text{fore}) \cdot \kappa_{\boldsymbol{\mathcal{H}}(s)}}$\Comment{optional step (Eq. \eqref{eq:scaling})}
            \State $\boldsymbol{\mathcal{X}}(s) \gets \mathcal{L}(\mathbf{t}_\text{fore}^{(z)}, \mathbf{X}_\text{fore}^{(z)})$ \Comment{by either Eq. \eqref{eq:DLT} or Eq. \eqref{eq:FFLT}}
            \State $\boldsymbol{\mathcal{Y}}(s) \gets \boldsymbol{\mathcal{H}}(s) (\mathbf{B}\boldsymbol{\mathcal{X}}(s) + \boldsymbol{\mathcal{P}}(s))$ \Comment{Eq. \eqref{eq:decomposed_LP}}
            \State \textbf{return} Numerical ILT $f_{\mathcal{L}^{-1}} (\boldsymbol{\mathcal{Y}}(s),\mathbf{t}_\text{fore})$ \Comment{by Eq. \eqref{eq:ILT_Fourier}}
        \EndFunction
        
        \State $\mathbf{Y}_\text{pred} \gets \emptyset$, $Q \gets \ceil[\bigg]{\dfrac{M}{\delta}}$
        \For{$q = 0$ to $Q-1$}
            \State $(\mathbf{t}_\text{fore}^{(z)}, \mathbf{X}_\text{fore}^{(z)}) \gets \text{Extract window}$
            \State $\mathbf{Y}_\text{pred}^{(z)} \gets \text{LPNet}(\mathbf{t}_\text{hist}, \mathbf{X}_\text{hist}, \mathbf{Y}_\text{hist}, \mathbf{t}_\text{fore}^{(z)}, \mathbf{X}_\text{fore}^{(z)}, N_{\mathrm{ILT}})$
            \State $(\mathbf{t}_\text{hist}, \mathbf{X}_\text{hist}, \mathbf{Y}_\text{hist}) \gets \text{Update history}$
            \State $\mathbf{Y}_\text{pred} \gets \mathbf{Y}_\text{pred} \cup \mathbf{Y}_\text{pred}^{(z)}$
        \EndFor
        
        \State \textbf{return} $\mathbf{Y}_\text{pred}$
    \end{algorithmic}
\end{algorithm}

\paragraph{External Input $X(s)$:} The Laplace transform of the external input can be computed numerically using either a discrete summation or a Fourier-based approach. The Discrete Laplace Transform (DLT) follows directly from the definition in Equation \eqref{eq:LT_analytic}:
\begin{equation}
\label{eq:DLT}
    \mathcal{X}(s) = \sum_{k=0}^{N-1} x(t_k)\,e^{-s t_k}\,\Delta t,
\end{equation}
where \(\Delta t = t_{k+1} - t_k\) is the time increment.

If \(x(t)\) is assumed periodic, it can be represented as a Fourier series \cite{cao2024}:
\begin{equation}\label{eq:fourier_excitaion}
    x(t) = \sum\limits_{k=-K}^{K} a_k e^{i\omega_k t}, \quad 0 \leq t < T,
\end{equation}
where \(a_k\) and \(\omega_k\) are the Fourier coefficients and frequencies, respectively. Using the Fast Fourier Transform (FFT), these coefficients can be efficiently computed. Applying the Laplace transform yields:
\begin{equation}
\label{eq:FFLT}
    \mathcal{X}(s) = \sum\limits_{k=-K}^{K} \frac{a_k}{s - i\omega_k}.
\end{equation}
This approach, referred to as the Fast Fourier Laplace Transform (FFLT), exploits the relationship between Fourier and Laplace transforms.

\paragraph{History Encoding and Initial State \(\boldsymbol{\mathcal{P}}(s)\)}
We represent the initial condition term \(\boldsymbol{\mathcal{P}}(s)\) via a combination of an encoder network and an analytic structure, rather than explicitly computing its Laplace transform. Specifically, we introduce a history‐dependent parameter \(\mathbf{P}\in\mathbb{R}^{D_y\times P}\) and a latent variable \(\mathbf{z}\), both inferred by a history encoder: 
\begin{equation}
  f_{\mathrm{enc}}\colon \mathbb{R}^{1} \times\mathbb{R}^{T_{\mathrm{hist}} \times d_x}\times \mathbb{R}^{T_{\mathrm{hist}} \times d_y}
  \;\to\;\mathbb{R}^{D_z \times P}\times\mathbb{R}^{D_z},
\end{equation}
such that:
\begin{equation}
\label{eq:hist_enc}
  \mathbf{P},\,\mathbf{z}
  \;=\;
  f_{\mathrm{enc}}\bigl(\mathbf{t_{\text{hist}}},\mathbf{X}_{\mathrm{hist}},\,\mathbf{Y}_{\mathrm{hist}}\bigr).
\end{equation}
Here, \(\mathbf{P}\) collects information needed to construct the polynomial representation of the initial state, and \(\mathbf{z}\) is a latent state capturing additional system characteristics from historical data.

Comparing \(\boldsymbol{\mathcal{P}}(s)\) to the analytic form in~\eqref{eq:P_func}, we note that \(\boldsymbol{\mathcal{P}}(s)\) is a polynomial in \(s\) whose coefficients depend on system matrices \(\mathbf{A}_i\) and initial states \(\tfrac{d^k \mathbf{y}(0)}{dt^k}\). Thus, the initial state term can be rewritten as:
\begin{equation}
\label{eq:init_Ps}
  P(s)
  \;=\;
  f_{\boldsymbol{\mathcal{P}}(s)}\!\bigl(s,\mathbf{P}\bigr)
  \;=\;
  \sum_{i=0}^{P-1}\mathbf{p}_i\,s^i
  \quad\text{with}\quad
  \mathbf{P}
  \;=\;
  \bigl(\mathbf{p}_0,\mathbf{p}_1,\dots,\mathbf{p}_{P-1}\bigr).
\end{equation}
This preserves the analytic dependence on initial conditions while offering a flexible, data‐driven adaptation. By encoding \(\mathbf{P}\) and \(\mathbf{z}\) jointly, the model can capture system properties without direct knowledge of \(\mathbf{A}_i\) or \(\mathbf{y}(0)\) and its derivates.

\paragraph{Transfer Function \(\boldsymbol{\mathcal{H}}(s)\)}
Next, we learn the overall transfer function \(\boldsymbol{\mathcal{H}}(s)\) via a NN $f_{\boldsymbol{\mathcal{H}}(s)}$, which takes as input a 2D-grid \(\mathbf{g} \in\mathbb{R}^{N_{\mathrm{ILT}}\times T}\). Here, \(N_{\mathrm{ILT}}\) is the number of ILT terms (see Eq.~\eqref{eq:ILT_Fourier}), and \(T\) is the forecast horizon. Although the ILT method naturally produces the (complex) query points \(s_k(t)\) (Eq.~\eqref{eq:queries}), these can grow unbounded for small \(t\), leading to extreme input values for the NN. This results in large weight updates, potentially causing gradient instability and slowing down convergence. To address this, we map the ILT queries onto a normalized grid \(\mathbf{g}=[g_{m,n}]\), ensuring numerical stability while preserving spectral and temporal structure. The ILT axis remains uniform, while irregularly sampled time points \( t_n \) are scaled to \([-1,1]\) using:  
\begin{equation}
\label{eq:grid}
    g_{m,n} = 2 \frac{t_n - t_{\min}}{t_{\max} - t_{\min}} - 1.
\end{equation}
This ensures that \( t_{\min} \) maps to \(-1\) and \( t_{\max} \) to \( 1 \), preserving relative spacing. 

The transfer function is then modelled by:
\begin{equation}
\label{eq:Hs_net}
  \boldsymbol{\mathcal{H}}(s) 
  \;=\;
  f_{\boldsymbol{\mathcal{H}}(s)}(\mathbf{g}, \mathbf{z}),
\end{equation}
where \(\mathbf{z}\) is precisely the latent state inferred by the history encoder \eqref{eq:hist_enc}. Conditioning on \(\mathbf{z}\) allows \(\boldsymbol{\mathcal{H}}(s)\) to adapt to varying system behaviours and capture locally linear approximations of potentially non-linear processes. 

\paragraph{Inverse Laplace Transform and Time-Independent Scaling}
In Eq.~\eqref{eq:ILT_Fourier}, the contour parameter is often set as \(\lambda = \zeta t\), making \(\boldsymbol{\mathcal{Y}}(s)\) explicitly time-dependent. To remove this dependency, we define the scaling factor:
\begin{equation}
\label{eq:scaling_factor}
  f_{\mathrm{scale}}(t) = \zeta t\, e^{-\sigma t} = \zeta t\, \epsilon^{\tfrac{1}{\zeta}} e^{-\alpha t}.
\end{equation}
When scaling is used, we get \(\widetilde{\boldsymbol{\mathcal{H}}}(s)\) by Eq. \eqref{eq:Hs_net} which is transformed into the time dependent transfer function along the $t$ axis by:
\begin{equation}
\label{eq:scaling}
  \boldsymbol{\mathcal{H}}(s) = \frac{\widetilde{\boldsymbol{\mathcal{H}}}(s)}{f_{\mathrm{scale}}(t) \cdot \kappa_{\boldsymbol{\mathcal{H}}(s)}}.
\end{equation}
Here, the scaling parameter $\kappa_{\boldsymbol{\mathcal{H}}(s)}$ stabilises training by preventing the amplification of small variations in \(\widetilde{\boldsymbol{\mathcal{H}}}(s)\) due to division by \( f_{\mathrm{scale}}(t) \), ensuring a well-conditioned and robust representation.
\section{Evaluation}
\label{sec:experiments}
We evaluate our method by benchmarking it against a sequence-to-sequence LSTM (seq2seq LSTM) and the LNO of \cite{cao2024} on eight univariate dynamical system datasets that differ in complexity and characteristic behaviour. By restricting ourselves to the univariate case, we concentrate on the core dynamics without the complexity of multiple input channels. In this formulation, the input-to-state $\mathbf{B}$ is simply a scalar, set to 1.

\paragraph{Experimental Setup.}
All datasets consist of uniformly sampled time series with 50 historical data points and a forecast horizon of 500 steps. We employ a train-validation-test split: the training set is used to learn model parameters, the validation set for hyperparameter tuning, and the test set for final performance. 

For the Spring-Mass-Damper System (SMD) and Mackey-Glass data, we generate our own univariate time series by applying three distinct periodic signals for training (sigmoid), validation (decaying sine), and testing (triangular). In contrast, for the Duffing, Lorenz, and driven pendulum datasets, we adopt the data of \cite{cao2024} directly, which rely on decaying sinusoidal input signals $x(t)$ whose decay coefficient varies across samples. To reduce experimental bias, we determine the learning rate and other model-specific hyperparameters via the Tree-Parzen Estimator (TPE) \cite{ozaki2020} in \texttt{optuna}, using 100 TPE trials per dataset. We then train each model six times with different random seeds to assess robustness. All datasets are processed in a single pass (batch size 512) per epoch. Our experiments run on a high-performance cluster with eight GPUs (Nvidia A40 and A100).

\paragraph{Test Systems and Results.}
Table~\ref{tab:residuals_test_Forecasting-Based} presents the mean and standard deviation of the MSE on the test set over six repeated runs. Below, we provide a brief overview of each system and comment on the observed results.
\begin{table*}[htb]
\centering
\caption{Mean ± (standard deviation) for MSE on test set. \textbf{Bold} values indicate the best result per data set.}
\label{tab:residuals_test_Forecasting-Based}
\begin{tabular}{llll}
\toprule
Dataset/Model & LNO & LSTM & LP-Net \\
\midrule
SMD & 1.88e-01 (1.48e-01) & \textbf{3.56e-04 (4.53e-05)} & 8.75e-04 (2.46e-04) \\
Duffing $c=0$ & 7.42e-02 (3.93e-02) & 1.21e-01 (2.21e-02) & \textbf{1.98e-02 (6.52e-03)} \\
Duffing $c=0.5$ & 1.06e-03 (1.04e-04) & 1.33e-03 (8.73e-04) & \textbf{5.31e-05 (1.23e-05)} \\
Lorenz $\rho=5$ & 4.05e-02 (1.18e-02) & \textbf{1.21e-04 (2.77e-05)} & 5.19e-04 (1.09e-04) \\
Lorenz $\rho=10$ & 5.36e+00 (5.97e-01) & 2.10e+00 (3.79e-01) & \textbf{1.31e+00 (3.27e-01)} \\
Pendulum $c=0$ & 5.81e-01 (9.83e-02) & 6.89e-01 (6.43e-02) & \textbf{6.61e-03 (1.67e-03)} \\
Pendulum $c=0.5$ & 1.08e-03 (7.37e-04) & 6.92e-04 (1.52e-04) & \textbf{5.10e-05 (2.00e-05)} \\
Mackey-Glass & 5.90e-01 (2.08e-01) & 3.50e-01 (6.99e-02) & \textbf{8.83e-03 (3.27e-03)} \\
\bottomrule
\end{tabular}

\end{table*}
We start with a simple linear \textbf{Spring-Mass-Damper (SMD) system:}
\begin{equation}
m\,\ddot{y}(t) + c\,\dot{y}(t) + k\,y(t) = x(t), 
\quad m,c,k \in \mathbb{R}^+,
\end{equation}
where \(m\) is the mass, \(c\) the damping, and \(k\) the spring constant. It has an initial displacement \(y(0) = y_0\) as well as initial velocity \(\dot{y}(0) = v_0\). As the table shows, the seq2seq LSTM achieves the lowest error, yet LP-Net still surpasses LNO and provides a stable fit. This indicates that, although SMD is comparatively simpler to model, LP-Net remains competitive.

The \textbf{Duffing oscillator:}
\begin{equation}
m\,\ddot{y}(t) + c\,\dot{y}(t) + k_1\,y(t) + k_3\,y^3(t) = x(t),
\quad m,c,k_1,k_3 \in \mathbb{R}^+,
\end{equation}
introduces a cubic spring (constant \(k_3\)). Following \cite{cao2024}, we use damped (\(c=0.5\)), which provides transient bahavior as the oscillations decline and undamped (\(c=0\)) where oscillatory behaviour dominated. Notably, LP-Net achieves the best results for both scenarios, capturing non-linear oscillatory behaviour accurately in the damped case.

\noindent
The \textbf{Lorenz system:} 
\begin{equation}
\dot{s}_x = \sigma (y - s_x), \quad
\dot{y} = s_x(\rho - s_z) - y, \quad
\dot{s}_z = s_x y - \beta s_z - x(t),
\quad \sigma,\rho,\beta \in \mathbb{R}^+,
\label{eq:Lorenz_system}
\end{equation}
is a three-dimensional chaotic model where $s_x,s_z$ are internal states, \(\rho=5\) or \(\rho=10\) sets the degree of chaos. For \(\rho=5\), the table indicates that the LSTM gives the best result. However, at \(\rho=10\), LP-Net emerges on top, evidencing superior adaptability under stronger chaotic dynamics.

\noindent
Another non-linear system is the \textbf{Driven pendulum:}
\begin{equation}
\ddot{x}(t) + c\,\dot{x}(t) + \frac{g}{l}\,\sin\bigl(x(t)\bigr) = x_{\text{ext}}(t),
\quad c,g,l \in \mathbb{R}^+,
\end{equation}
where $g$ is the gravity and $l$ is the rod length. Again a damped (\(c=0.5\)) and undamped (\(c=0\)) similar to the Duffing system, a higher damping leads to more transient, decaying behaviour. Here, LP-Net clearly outperforms LSTM and LNO in both scenarios, underscoring its strength in modelling non-linear oscillatory phenomena.

\noindent
The \textbf{Mackey-Glass system:}
\begin{equation}
\dot{y}(t) = \beta \frac{y(t-\tau)}{1 + [y(t-\tau)]^n} 
           - \gamma y(t) + x(t),
\quad \beta,\gamma,\tau,n \in \mathbb{R}^+,
\end{equation}
brings explicit time delays $\tau$, which often pose challenges for standard recurrent architectures. Here, \( \beta \) controls the strength of the delayed feedback, while \( \gamma \) represents the dissipation rate. LP-Net attains a particularly low error, whereas for the LSTM and LNO we observe that they are not able to capture the behaviour. Because all models share the same training and tuning procedures, this improvement suggests LP-Net retains more effective long-range memory, thus handling delayed feedback more robustly.

Taken together, these findings verify that LP-Net is accurate and robust across diverse types of dynamical systems. In most cases, it surpasses both seq2seq LSTM and LNO, particularly for non-linear, chaotic, or delay-based dynamics. This ability to handle forced, damping-induced, and chaotic regimes demonstrates the method’s versatility for extended forecasting horizons.
Table \ref{tab:residuals_test_Forecasting-Based} shows that our method, is able to outperform LNO as well as the LSTM for most of the datasets, except for the Lorenz System with $\rho=5$.

\section{Discussion and Limitations}  
\label{sec:Discussion}  

While LP-Net offers a flexible, solver-free approach to modelling forced dynamical systems, it has theoretical and practical limitations. A fundamental constraint arises from the Laplace transform: functions growing super-exponentially, i.e., those for which there exist no constants \( C, \sigma > 0 \) such that \( |f(t)| \leq C e^{\sigma t} \), are not transformable \cite{schiff1999}. This limitation is particularly relevant for transforming input signals $\boldsymbol{\mathcal{X}}(s)=\mathcal{L}\{\mathbf{x}(t)\}$, although in practice, such growth is rare in real-world applications.

Another challenge lies in the numerical approximation of \(\boldsymbol{\mathcal{X}}(s)\) and \(\boldsymbol{\mathcal{Y}}(s)\). For signals containing high frequencies (e.g. abrupt jumps), a large number of ILT terms (\(N_{\text{ILT}}\)) is required for accurate reconstruction. While the computation of the proposed Laplace transforms, such as the Discrete Laplace Transform (DLT) in Eq. \eqref{eq:DLT} and the Fast Fourier Laplace Transform (FFLT) in Eq. \eqref{eq:FFLT}, can be computationally intensive, it can be performed once prior training.

Memory consumption also presents a constraint, as the computational cost grows linearly with both the number of ILT terms \(N_{\text{ILT}}\) and the number of prediction time points. Particularly for signals with sharp discontinuities, achieving sufficient accuracy requires large \(N_{\text{ILT}}\) values, which significantly increases memory requirements. 
Despite these challenges, the design of LP-Net is highly parallelisable, with the exception of time-stepping. However, this overhead is negligible compared to the numerous iterations required by ODE solvers.

One major advantage of LP-Net is that the Laplace representation of the input $\boldsymbol{\mathcal{X}}(s)$ is handled independently, allowing for pre-validation or even manual refinement before joint training. This makes debugging and refining the forcing component significantly easier. Moreover, if certain components of the system transfer function $\boldsymbol{\mathcal{H}}(s)$ are already known, for instance, from physical principles, they can be directly embedded into the model while learning only the remaining unknown terms. This hybrid approach reduces the learning burden and allows for more effective integration of prior knowledge.

The decoupled structure also enables selective training strategies. When working with controlled experiments where the system is initialized at zero, the initial state term $\boldsymbol{\mathcal{P}}(s)$ can be explicitly set to zero, removing unnecessary degrees of freedom and simplifying optimization. In real-world settings where the initial conditions vary, the full model can then be fine-tuned, leveraging pre-trained components for efficient adaptation.

Furthermore the decoupled structure of LP-Net improves interpretability: In many fields, the structure of the dynamical system, learned with $\boldsymbol{\mathcal{H}}(s)$ is of interest (e.g. when designing the controller $C$). Learning it as a dedicated component enables domain experts to interpret what was learned.

\section{Conclusion and Future Work}
\label{sec:Conclusion}
In this work, we introduced a decoupled, solver-free neural network (NN) framework for learning forced and delay-aware dynamical systems, the Laplace-based Network (LP-Net). LP-Net explicitly separates internal system dynamics from external control inputs and initial conditions, addressing limitations of current literature approaches. 
Thus, LP-Net enhances transferability, interpretability, scalability: rapid retraining or fine-tuning for new forcing signals, offering a significant advantage in scenarios where data is limited; the modules mimic classical theoretical concepts known by experts (e.g. transfer functions); highly parallelisable and scales linearly with the number of frequency terms and prediction time-steps, respectively.

LP-Net demonstrates clear advantages across a wide range of dynamical systems. Evaluated on eight datasets covering linear, non-linear, chaotic, and delayed dynamics, it consistently outperforms the Laplace Neural Operator (LNO) in all cases. Compared to an Long Short-Term Memory (LSTM) model, LP-Net achieves superior accuracy on nearly all datasets, with the exception of a linear ODE case and one out of seven non-linear scenarios, as shown in Table \ref{tab:residuals_test_Forecasting-Based}. These results highlight the effectiveness of LP-Net in capturing complex system behaviour across various dynamical systems.

Several directions for future work can be taken: investigate the performance of using more specialised NN architectures for the components of LP-Net beyond fully-connected networks; evaluate on real-world datasets and multi-variate time-series; extend to handle 2 and 3 dimensional use cases, similar to LNO.

\bibliographystyle{splncs04}
%\bibliography{main}

\newpage
\appendix
\begin{center}
{\Huge Supplementary Material}
\end{center}
\setcounter{page}{1}

\section{Appendix: Derivation of the Time-Dependent Scaling Factor}
\label{sec:app_ilt_scaling}

We begin with the ILT approximation given by
\begin{equation}\label{eq:ILT_Fourier_appendix}
    \mathbf{y}(t) \approx \frac{1}{\lambda} e^{\sigma t} \biggl[
    \frac{\mathbf{Y}(\sigma)}{2}
    + \sum_{k=1}^{N_{\text{ILT}}}
    \mathrm{Re} \biggl\{
      \mathbf{Y} \Bigl(\sigma + i\,\frac{k\pi}{\lambda} \Bigr)
      e^{i \frac{k\pi t}{\lambda}}
    \biggr\}
    \biggr],
\end{equation}
where \(\sigma = \alpha - \frac{\log(\epsilon)}{\lambda}\). In many implementations, \(\lambda\) is chosen proportional to \(t\), i.e.,  
\begin{equation}
    \lambda = \zeta t.
\end{equation}
Hence,  
\begin{equation}
    \sigma = \alpha - \frac{\log(\epsilon)}{\zeta t}
    \quad\Longrightarrow\quad
    \sigma t = \alpha t - \frac{\log(\epsilon)}{\zeta}.
\end{equation}
It follows that
\begin{equation}
    e^{\sigma t} = e^{\alpha t} \epsilon^{-\frac{1}{\zeta}}
    \quad\text{and}\quad
    \frac{1}{\lambda} e^{\sigma t}
    = \frac{1}{\zeta t} e^{\alpha t} \epsilon^{-\frac{1}{\zeta}}.
\end{equation}
For large \(t\), this term can become numerically unstable in the ILT formula. To remove it, we define a \emph{time-dependent} scaling:
\begin{equation}
    f_{\mathrm{scale}}(t)
    = \lambda e^{-\sigma t}
    = \zeta t e^{-\alpha t} \epsilon^{\frac{1}{\zeta}}.
\end{equation}
Now introduce a \emph{scaled} function:
\begin{equation}
    \widetilde{\mathbf{Y}}(s)
    = f_{\mathrm{scale}}(t) \mathbf{Y}(s),
\end{equation}
so that the product \(\frac{1}{\lambda} e^{\sigma t} \mathbf{Y}(s)\) in \eqref{eq:ILT_Fourier_appendix} simplifies to \(\widetilde{\mathbf{Y}}(s)\) inside the brackets.  
Thus, \(\mathbf{y}(t)\) can be rewritten \emph{without} a large prefactor multiplying \(\mathbf{Y}(s)\):  
\begin{equation}
    \mathbf{y}(t) \approx
    \biggl[
    \frac{\widetilde{\mathbf{Y}}(\sigma)}{2}
    + \sum_{k=1}^{N_{\text{ILT}}}
    \mathrm{Re} \Bigl\{
      \widetilde{\mathbf{Y}} \Bigl(\sigma + i\,\frac{k\pi}{\lambda} \Bigr)
      e^{i \frac{k\pi t}{\lambda}}
    \Bigr\}
    \biggr].
\end{equation}
Thus, scaling \(\mathbf{Y}(s)\) by \(f_{\mathrm{scale}}(t)\) removes the exponential growth or decay induced by \(\sigma t\) and the factor \(\frac{1}{\lambda}\), resulting in a more numerically stable representation for \(\mathbf{y}(t)\).
\newpage
\section{Details on the Evaluation}
\subsection{Models}
The following section provides a description of the hyper parameters used in the dieefent models
\subsubsection{Laplace-Net}
\begin{itemize}
    \item \textbf{$LP_{type}$}: Specifies the type of forward Laplace transform used to transform $\mathbf{x}(t)$ into $\boldsymbol{\mathcal{X}}(s)$. DLT stands for Direct Laplace Transform (Eq. \eqref{eq:DLT}), and FFLT stands for Fast Fourier Laplace Transform (Eq. \eqref{eq:FFLT}).
    \item \textbf{$\alpha$}: Positive shift of the real axis for the queries and the ILT (Eq. \eqref{eq:queries}).
    \item \textbf{$\zeta$}: Scaling parameter for the ILT, defined as $\lambda=\zeta t$ in Eq. \eqref{eq:ILT_Fourier} and Eq. \eqref{eq:queries}.
    \item \textbf{c-shift}: Contour shift along the time axis. The $\mathbf{t}_\text{fore}$ time axis is shifted in the positive direction to avoid small $t$ values.
    \item \textbf{$N_{ILT}$}: Number of terms for the ILT (Eq. \eqref{eq:ILT_Fourier}).
    \item \textbf{$d_{Enc}$}: Dimensionality of the encoder network, determining the number of hidden units in its layers.
    \item \textbf{$L_{Enc}$}: Number of layers in the encoder network.
    \item \textbf{P}: Assumed degree of the differential equation, used to determine the number of polynomial terms in Eq. \eqref{eq:P_func}.
    \item \textbf{$\kappa_{\boldsymbol{\mathcal{H}}(s)}$}: A scaling factor applied to the Laplace-transformed representation.
    \item \textbf{lr}: Learning rate for model training, scaled by $10^{-3}$.
    \item \textbf{Q}: Number of recurrent steps. While Algorithm \ref{alg:lpnet} determines them based on the stride $\delta$, we explicitly define the number of steps in our experiments.
    \item \textbf{$Act_{H(s)}$}: Activation function used in the transfer function network $f_{\boldsymbol{\mathcal{H}}(s)}(\mathbf{g}, \mathbf{z})$.
    \item \textbf{$d_{H(s)}$}: Dimensionality of the hidden layers of $f_{\boldsymbol{\mathcal{H}}(s)}(\mathbf{g}, \mathbf{z})$.
    \item \textbf{$l_{H(s)}$}: Number of layers used in $f_{\boldsymbol{\mathcal{H}}(s)}(\mathbf{g}, \mathbf{z})$.
\end{itemize}

\subsubsection{LSTM}
\begin{itemize}
    \item \textbf{Hidden Dim}: Number of hidden units in each LSTM layer.
    \item \textbf{lr}: Learning rate for model training, scaled by $10^{-3}$.
    \item \textbf{$L$}: Number of LSTM layers stacked in the sequence-to-sequence model.
\end{itemize}

\subsubsection{LNO}
\begin{itemize}
    \item \textbf{width}: Defines the number of channels in each layer, determining the feature representation size in the model.
    \item \textbf{modes}: Specifies the number of frequency components used in the Laplace transform, affecting the resolution of the learned representation.
\end{itemize}

\subsection{Datasets}
Table \ref{tab:dataset_numbers} summarises the hyper parameter that where used by us or \cite{cao2024} to generate the datasets used for the experiments. All datasets start with zero initial conditions except for the Lorenz system where $S_x(0)=1$. The DS column refers the dataset numbers for the following tables.
\begin{table}[h]
    \centering
    \caption{Dataset numbering used in hyperparameter search.}
    \label{tab:dataset_numbers}
    \begin{tabular}{l|l|l|c|l}
        \hline
        \textbf{DS} & \textbf{Dataset Name} & \textbf{Parameters} & \makecell{\textbf{Samples} \\ \textbf{train-val-test}} & $T$ (sek)\\
        \hline
        1 & SMD  & $m=1$, $c=0.5$, $k=5$ & 10 - 5 - 15 &20 \\
        2 & Duffing $c=0$ &$m=1$, $c=0$, $k_1=1$, $k_3=1$ &  200-50-130& 20.47 \\
        3 & Duffing $c=0.5$ &$m=1$, $c=0.5$, $k_1=1$, $k_3=1$  & 200-50-130&20.47 \\
        4 & Lorenz $\rho=5$ &$\sigma=10$,$\beta=8/3$,$\rho=5$ &  200-50-130&20.47 \\
        5 & Lorenz $\rho=10$  &$\sigma=10$,$\beta=8/3$,$\rho=10$ & 200-50-130&20.47 \\
        6 & Pendulum $c=0$ & $g/l = 1$,$c=0$ &  200-50-130&20.47\\
        7 & Pendulum $c=0.5$ &$g/l = 1$,$c=0.5$  & 200-50-130&20.47 \\
        8 & Mackey-Glass & $\beta=0.1$, $\gamma=0.2$, $\tau=7$, $n=2$ & 10 - 5 - 5&20 \\
        \hline
    \end{tabular}
\end{table}

\subsection{Hyperparameters}
We summarize the hyperparameters obtained through automated hyperparameter tuning, optimized based on the validation loss using the TPE algorithm.  

The dataset indices in DS correspond to the datasets listed in Table \ref{tab:dataset_numbers}.  
Table \ref{tab:hparams_LPNet} presents the identified hyperparameters for LP-Net, while Table \ref{tab:seq2seq_lstm2} provides the hyperparameters for the LSTM model. The hyperparameters for the Laplace Neural Operator are shown in Table \ref{tab:seq2seq_LNO}.  

\begin{table*}[h]
\caption{Hyperparameters for LP-Net}
\label{tab:hparams_LPNet}
\begin{tabular}{l|l|l|l|l|l|l|l|l|l|l|l|l|l|l}
\toprule
 DS& $LP_{type}$ & $\alpha 10^{-3}$& $\zeta$ & c- shift & $N_{ILT}$ & $d_{Enc}$ & $L_{Enc}$ & P & $\kappa_{\boldsymbol{\mathcal{H}}(s)}$ & lr$10^{-3}$ & $Q$ & $Act_{H(s)}$& $d_{H(s)}$ & $l_{H(s)}$ \\
\midrule
1 & DLT & 4.51 & 2.0 & 2.7 & 41 & 56 & 2 & 3 & 450 & 4.40 & 3 & tanh & 192 & 4 \\
2 & DLT & 2.26 & 2.7 & 1.5 & 37 & 64 & 4 & 1 & 330 & 1.86 & 2 & tanh & 144 & 2 \\
3 & FFLT & 9.81 & 2.5 & 4.5 & 41 & 40 & 2 & 3 & 270 & 3.98 & 4 & softsign & 96 & 2 \\
4 & DLT & 6.46 & 1.5 & 4.4 & 43 & 24 & 1 & 2 & 100 & 3.73 & 5 & silu & 144 & 4 \\
5 & FFLT & 3.46 & 2.7 & 7.4 & 67 & 48 & 1 & 3 & 440 & 0.74 & 2 & silu & 160 & 6 \\
6 & FFLT & 2.81 & 2.1 & 4.3 & 67 & 56 & 1 & 2 & 380 & 2.5 & 1 & silu & 64 & 6 \\
7 & FFLT & 9.71 & 1.7 & 8.2 & 75 & 20 & 5 & 3 & 400 & 3.95 & 8 & softsign & 48 & 6 \\
8 & FFLT & 7.26 & 2.6 & 9.6 & 79 & 16 & 2 & 3 & 110 & 5.88 & 10 & silu & 64 & 1 \\
\bottomrule
\end{tabular}
\end{table*}

\begin{table}[h]
    \centering
    \caption{Hyperparameters for the Seq2Seq LSTM model.}
    \label{tab:seq2seq_lstm2}
    \begin{tabular}{c r r r}
        \toprule
        DS & Hidden Dim & lr $10^{-3}$ & $L$ \\
        \midrule
        1 & 488 & 0.03 & 1 \\
        2 & 184 & 0.85 & 5 \\
        3 & 192 & 0.42 & 1 \\
        4 & 344 & 1.08 & 1 \\
        5 & 112 & 3.87 & 1 \\
        6 & 168 & 4.31 & 4 \\
        7 & 280 & 2.35 & 2 \\
        8 & 144 & 0.23 & 4 \\
        \bottomrule
    \end{tabular}
\end{table}  

\begin{table}[h]
    \centering
    \caption{Hyperparameters for the Laplace Neural Operator model.}
    \label{tab:seq2seq_LNO}
    \begin{tabular}{c l r r r}
        \toprule
        DS & Act. & lr $10^{-3}$ & Modes & Width \\
        \midrule
        1 & tanh & 0.568 & 25 & 4 \\
        2 & sin  & 1.060 & 53 & 4 \\
        3 & sin  & 0.364 & 17 & 4 \\
        4 & sin  & 3.033 & 16 & 4 \\
        5 & tanh & 1.263 & 54 & 4 \\
        6 & tanh & 9.469 & 36 & 2 \\
        7 & sin  & 1.887 & 47 & 4 \\
        8 & tanh & 0.104 & 35 & 4 \\
        \bottomrule
    \end{tabular}
\end{table}

\end{document}